\documentclass[english]{article}
\usepackage[T1]{fontenc}
\usepackage[latin9]{inputenc}
\usepackage{babel}
\usepackage{url}
\usepackage{graphicx}
\usepackage[unicode=true]
 {hyperref}

\makeatletter
\@ifundefined{showcaptionsetup}{}{%
 \PassOptionsToPackage{caption=false}{subfig}}
\usepackage{subfig}
\makeatother

\begin{document}

\title{Simion Zoo: A Workbench for Distributed Experimentation with Reinforcement
Learning for Continuous Control Tasks}

\author{Borja Fernandez-Gauna, Manuel Gra\~na and Roland S. Zimmermann}
\maketitle
\begin{abstract}
We present Simion Zoo, a Reinforcement Learning (RL) workbench that
provides a complete set of tools to design, run, and analyze the results,both
statistically and visually, of RL control applications. The main features
that set apart Simion Zoo from similar software packages are its easy-to-use
GUI, its support for distributed execution including deployment over
graphics processing units (GPUs) , and the possibility to explore
concurrently the RL metaparameter space , which is key to successful
RL experimentation.
\end{abstract}

\section{Introduction\label{sec:Introduction}}

In recent years, Reinforcement Learning (RL) has become a very popular
area of research, because of the almost exponential increase in computing
power  due to the advent of dedicated GPUs that have empowered researchers
to face previously unaffordable problems. In particular, the successful
applications of Deep Reinforcement Learning (DRL)to produce master videogame
players \cite{Vinyals2017,Silver2016} have created  great expectations
about the potential of DRL, even outside the academic research community.
As a result of this popularity boost, the number of RL software packages
has grown significantly. Nevertheless, these projects are mostly oriented
towards the research community, i.e. they assume sophisticated programming
users with  powerful computing resources to run the software. Even
for sophisticated programmers, these packages impose a steep learning
curve that hinders their user experience. This is in stark contrast
with the \emph{de-facto} user standards for\emph{Supervised Learning}
(SL) software, which customarily allow users to design/run experiments,
and to analyze the results on an intuitive Graphical User Interface
(GUI) that allows a swift learning curve. Users without programming
skills that intend to design and run RL experiments quickly on inexpensive
and commonly available hardware will obviously appreciate such kind
of facilities. Let use numerate the most important features that  a
user-friendly tool should possess: (a) easy installation, (b) a GUI
usable by non-programmers, (c) graphical visualization of the experiment,
and relevant data structures, which helps to understand the performance
achieved by the learning process, (d) facilities for statistical analysis
of the results, (e)concurrent exploration of metaparameter space shortening
the trial-and-error cycle, (f) efficient use of heterogeneous computing
resources, and (g) easy application to a broad range of RL problems,
algorithms and controllers.

The paper is structured as follows: first, Section \ref{sec:Simion-Zoo-Framework}
presents the Simion Zoo workbench, Section \ref{sec:Related-Work}
briefly reviews related works, and Section \ref{sec:Documentation,-Requirements-and}
discusses the availability of the project documentation  and its requirements.

\section{Simion Zoo Workbench\label{sec:Simion-Zoo-Framework}}

\emph{Simion Zoo \cite{borja_fernandez_gauna_2019_2579013}} is a
workbench to testRL and DRL algorithms that focuses on model-free
RL learning algorithms applied to control problems defined on continuous
state and action spaces. This software was designed to fulfill the
requirements enumerated in Section \ref{sec:Introduction}. (a) The
installation process is  straight-forward. One single installer is
provided to install the \emph{Herd Agent} service/daemon on the slave
machines. On the master computer, the user needs only to run \emph{Badger}
(the main application) which requires no installation and bundles
all the dependencies. (b) The configuration, execution and analysis
of results of the RL experiments is done via a user-friendly GUI.
(c) Experiments can  be run either locally in the master computer or
distributed over the slave computers. Locally run experiments can
be visualized live, whereas remotely executed experiments are monitorized
live (showing the average episode rewards) but can also be visualized
off-line once finished. (d) Finished experiments can be further analyzed
with a provided tool that generates publication-quality plots and statistics
of the system variables. (e) Parameters can be \emph{forked} and given
as many values as desired, so that experiments with all the parameter
value combinations are run concurrently. (f) Each slave machine receives binaries
which are compatible with its own operating system and architecture,
taking advantage of all the resources available on the computer (all
the CPU cores and/or the GPU). The project currently supports \emph{Windows}
and \emph{Linux} operating systems. (g) The workbench features a wide
set of built-in environments and agents.

The user can use conventional controllers (Proportional-Integrative-Derivative,
Linear-Quadratic Regulator, Variable-Speed Wind-Turbine (VSWT)), specific
controllers \cite{Fernandez-Gauna2017}, Q-function learning algorithms
(SARSA, Q-Learning, and Double Q-Learning \cite{Hasselt2010}), Actor-Critic
algorithms (CACLA, regular gradient ascent, Incremental Natural Actor-Critic
\cite{Degris2012}, Off-Policy Actor-Critic \cite{Degris2012a}, and
Off-Policy Deterministic Actor-Critic \cite{Silver2014}), and Deep
RL methods (DQN, Double-DQN and DDPG). Besides, policy learners (Actors)
can be combined with value function learners (Critics): Temporal-Difference$\left(\lambda\right)$,
TDC$\left(\lambda\right)$, and True Online Temporal-Difference \cite{Seijen2016}.

The workbench offers a broad set of built-in simulation environments:
classical benchmarking control tasks (mountain-car, balancing pole,
swing-up pendulum, and double swing-up pendulum), some benchmark tasks
from \cite{Hafner2011} (underwater vehicle control and airplane pitch
control), several single and multi-robot control problems that use
the \emph{Bullet Physics} library, and two VSWT models (a two-mass
model, and \emph{OpenFAST}\footnote{\emph{https://nwtc.nrel.gov/FAST}},
which is considered the state-of-the-art of Wind Turbine simulation).

\subsection{Using Simion Zoo\label{sec:Using-SimionZoo}}

The main application is \emph{Badger}, which offers a three-phase
experimentation process pipeline: design, monitor and analysis. Each
 phase has a dedicated tab within the \emph{Badger}'sGUI. In the \emph{Editor}
tab, the user selects the agent type,the simulation environment, and
sets the value/options of their  parameters. The learning and simulation
parameters, aka metaparameters, are organized in a hierarchy, they
change dynamically depending on the user choices, and they can be
given several values. Once the experiments are designed (we may  design
and run several experiments concurrently), the user can press the
\emph{Launch} button generating a set of experimental instances units,
each  corresponding to a combination of metaparameter values , and
 switch afterwards to the \emph{Monitor} tab. This tab shows on its
left part a list of the available agents and their capabilities, allowing the
user to select all or a subset of them for the next experiment. Once
started, the progress of each experimental unit is shown, allowing
the user early experiment cancellation  if the learning performance
is not as good as expected. Finally, the user may analyze the learning
and simulation results in the \emph{Reports} tab, selecting a subset
of the logged variables, grouping experimental units by parameter
value, and visualizing the experiment\footnote{The user guide can be accessed online: \href{https://github.com/simionsoft/SimionZoo/wiki/User-guide}{https://github.com/simionsoft/SimionZoo/wiki/User-guide}}.
Most remarkably, SimionZoo generates customizable plots ready for
publication such as those published in \emph{\cite{Fernandez-Gauna2017}.}

\begin{figure}
\begin{centering}
\subfloat{\begin{centering}
\includegraphics[width=0.45\columnwidth]{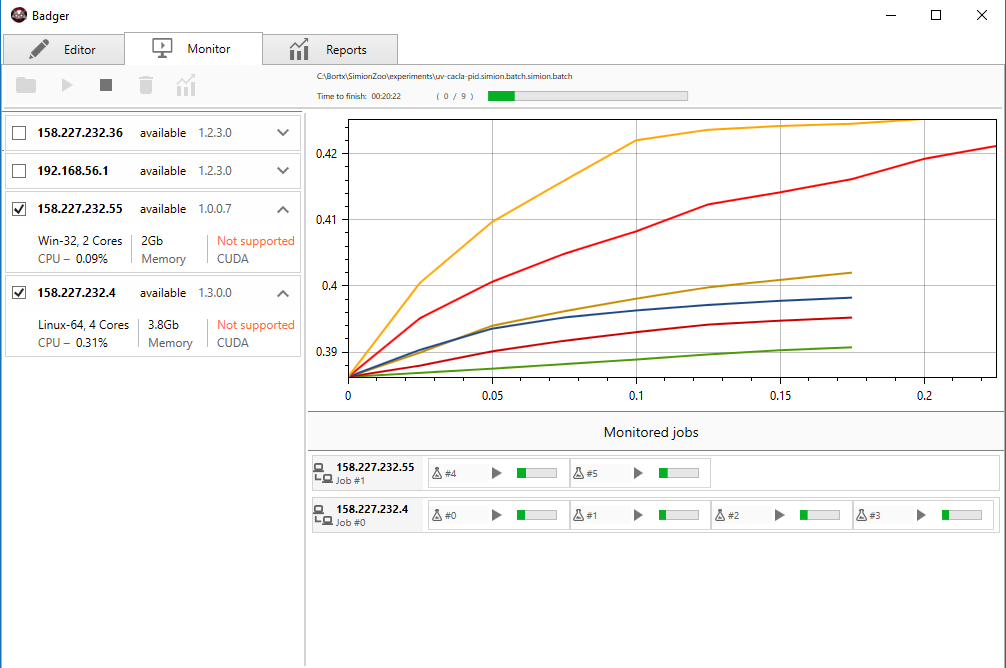}
\par\end{centering}
\label{badger-monitor}}\subfloat{\begin{centering}
\includegraphics[width=0.4\columnwidth]{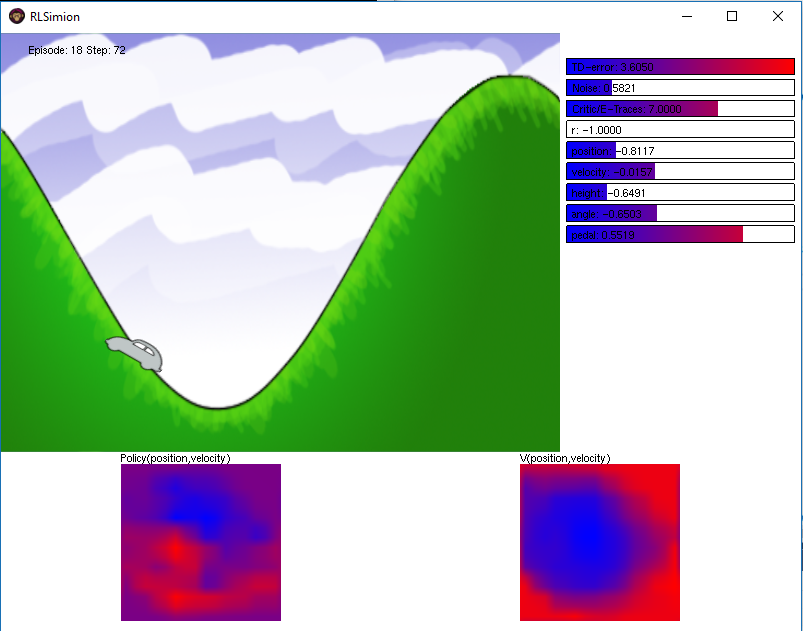}
\par\end{centering}
\label{visualization-mountain-car}}
\par\end{centering}
\caption{Two screenshots of \emph{Badger}: on the left, the \emph{Monitor}
tab and, on the right, a visualization of the Mountain-car environment.}

\end{figure}

\subsection{Extending Simion Zoo\label{subsec:Extending-Simion-Zoo}}

Extension or modification of the source code does not require to change
the GUI. The source code is automatically parsed after compilation
to generate the object class definitions and their parameters so that
the GUI  automatically adapts to the latest version of the code\footnote{The developer guide can be accessed online: https://github.com/simionsoft/SimionZoo/wiki/Developer-guide}.

\section{Related Work\label{sec:Related-Work}}

There are several programming libraries that provide RL-related functionalities,
i.e. Deep Neural Network libraries such as \emph{Tensor Flow}\footnote{https://www.tensorflow.org/},
\emph{Caffe}\footnote{http://caffe.berkeleyvision.org/},\emph{ }or
\emph{Microsoft Cognitive Toolkit}\footnote{https://www.microsoft.com/en-us/cognitive-toolkit/}.
These libraries provide the bottom layer to buildDRL algorithms.. Above
in the hierarchy are RL libraries that provide algorithm implementations
and environments but no GUIs or a configurable executable. Some of
the most popular are \emph{pyBrain} \footnote{http://www.pybrain.org/},
\emph{RL Park}\footnote{http://rlpark.github.io/site}, \emph{RLLib}\footnote{https://github.com/samindaa/RLLib},
and \emph{RL Library}\footnote{http://library.rl-community.org}.
Closest to our RL workbench software, we note two RL simulation environments
that offer a full GUI to edit, run and view/analyze RL experiments:
\emph{Maja Machine Learning Framework}\footnote{http://mmlf.sourceforge.net}
(\emph{MMLF}) and \emph{RL Sim}\footnote{https://www.cs.cmu.edu/\textasciitilde awm/rlsim/}.
The former does not support distributed executions, GPUs or multi-thread
execution, and seems to have been abandoned. The latter is a very
simple educational tool with only one configurable grid world environment
and some of the most typical tabular RL algorithms.

\section{Documentation, Licensing and Availability\label{sec:Documentation,-Requirements-and}}

SiminZoo has been published as open source in Zenodo \cite{borja_fernandez_gauna_2019_2579013}.Contributions
to the project's source code can be made through our public Github
repository (\url{https://github.com/simionsoft/SimionZoo}), where
the reader can also find the documentation (\url{https://github.com/simionsoft/SimionZoo/wiki}\emph{)}
and pre-compiled Windows and Linux binaries for the end-user (\url{https://github.com/borjafdezgauna/SimionZoo/releases}).
\emph{Simion Zoo} currently works under \emph{Windows} and \emph{Linux},
and is licensed under an \emph{MIT} license.

\section{Acknowledgements}

This project has received funding from the European Union's Horizon
2020 research and innovation programme under the Marie Sklodowska-Curie
grant agreement No 777720. The work reported in this paper has been
partially supported by FEDER funds for the MINECO project TIN2017-85827-P,
and projects KK-2018/00071 and KK-2018/00082 of the Elkartek 2018
funding program of the Basque Government. We would like to thank Unai
Tercero, Asier Rodriguez-Gonzalez and José-Alejandro Guerra-Denis
for their contributions to the project.

\bibliographystyle{plain}
\bibliography{ReinforcementLearning,FeedbackControl}

\end{document}